\documentclass[review]{elsarticle}
\usepackage{amssymb}
\usepackage{amsmath}
\usepackage{graphicx}
\usepackage{algorithm}
\usepackage{algpseudocode}
\usepackage{lineno}

\algnewcommand\algorithmicinput{\textbf{Input:}}
\algnewcommand\Input{\item[\algorithmicinput]}
\algnewcommand\algorithmicoutput{\textbf{Output:}}
\algnewcommand\Output{\item[\algorithmicoutput]}

\begin{document}
	
\ifpreprint
\setcounter{page}{1}
\else
\setcounter{page}{1}
\fi

\begin{frontmatter}
			
\title{Reorganizing local image features with chaotic maps: an application to texture recognition}

\author[imecc]{Joao B. Florindo\corref{cor1}} 
\cortext[cor1]{Corresponding author}
\ead{florindo@unicamp.br}

\address[imecc]{Institute of Mathematics, Statistics and Scientific Computing - University of Campinas\\
	Rua S\'{e}rgio Buarque de Holanda, 651, Cidade Universit\'{a}ria "Zeferino Vaz" - Distr. Bar\~{a}o Geraldo, CEP 13083-859, Campinas, SP, Brasil}

\begin{abstract}
Despite the recent success of convolutional neural networks in texture recognition, model-based descriptors are still competitive, especially when we do not have access to large amounts of annotated data for training and the interpretation of the model is an important issue. Among the model-based approaches, fractal geometry has been one of the most popular, especially in biological applications. Nevertheless, fractals are part of a much broader family of models, which are the non-linear operators, studied in chaos theory. In this context, we propose here a chaos-based local descriptor for texture recognition. More specifically, we map the image into the three-dimensional Euclidean space, iterate a chaotic map over this three-dimensional structure and convert it back to the original image. From such chaos-transformed image at each iteration we collect local descriptors (here we use local binary patters) and those descriptors compose the feature representation of the texture. The performance of our method was verified on the classification of benchmark databases and in the identification of Brazilian plant species based on the texture of the leaf surface. The achieved results confirmed our expectation of a competitive performance, even when compared with some learning-based modern approaches in the literature.
\end{abstract}


\end{frontmatter}

\section{Introduction}

Texture recognition has been one of the most important tasks in computer vision for many years. Examples of applications are found in medicine \cite{LZLSL18}, botany \cite{TF20}, geology \cite{NAHW20}, surveillance \cite{ALA20}, etc.

Despite the recent success of deep learning approaches in this area, like convolutional neural networks \cite{G16}, we still have space for research on model-based algorithms, especially when there is no large amount of annotated data available for training and the interpretation of the model outcome is relevant to the real domain of the problem. Such scenario frequently happens in areas like medicine, for example, where annotation by human specialists is laborious and expensive, and the interpretation of the machine verdict is fundamental to formulate reproducible hypotheses with theoretical foundations in medicine.

When it comes to textures, most of the well-succeeded hand-engineered approaches can be roughly divided into two categories: the local statistics descriptors, such as bag-of-visual-words \cite{VZ09} and local binary patterns \cite{OPM02}, and the model-based methods like fractal dimension \cite{XJF09}, jet spaces \cite{TG12} and complex networks \cite{GSFB16}. Whilst the first category usually comprises algorithms with high accuracy even using low computational resources, the second one provides interpretability as they are associated to physical characteristics of the material represented in the image rather than only relying on pixel distribution. Such complementary aspect of each approach ends up being a strong suggestions for hybrid approaches, combining both local statistics and physical modeling.

One of such model-based approaches that have been investigated for texture recognition, especially in biological sciences, is fractal geometry. However, fractals are part of a broader research area, which comprises non-linear dynamics and chaos theory. More general tools in these areas have been comparatively little explored for pattern recognition, despite the well established success of chaos-based techniques, for example, in data encryption \cite{JK01}.

Taking into consideration this background, we propose here a hybrid algorithm for texture classification combining the well established idea of local texture descriptors with one of the most popular tools in chaos theory, namely, chaotic maps \cite{G87}. Although this tool has been extensively studied in crypto\-graphy, its use in recognition is quite rare. And the reason for that is somewhat obvious: the primary purpose of a chaos map is to disorganize patterns, whereas recognition algorithms are supposed to rely on those patterns in their native condition to perform prediction. Nevertheless, we demonstrate here that the controlled use of those maps can be beneficial to recognition, inasmuch as such rearranging of patterns gives rise to complex non-linear relations that would not be learned by the algorithm in its classical configuration.

In more details, we propose the representation of a gray-scale image in a three-dimensional normalized Euclidean space and apply a chaotic map over this representation of the original image. The result after each iteration of the map is remapped to the image space and a distribution of local descriptors is computed to compose the representative feature vector. Several iterations of different maps are evaluated and we also consider intermediate steps between two successive map iterations. For that, we take a weighted combination between the remapped image in the previous and posterior iterations. Here we employ local binary patterns \cite{OPM02} as local descriptors, even though in principle other similar descriptors could be used in similar way.

The performance of the proposed method is assessed in terms of accuracy on the classification of texture databases whose results provided by other solutions has been extensively discussed in the literature. More exactly, we show results on KTHTIPS-2b \cite{HCFE04}, UIUC \cite{LSP05}, UMD \cite{XJF09}, and in an application to the identification of Brazilian plants based on the leaf surface (database 1200Tex) \cite{CMB09}. The achieved results confirmed our expectations of a competitive method, even when compared to some learning-based solutions in the literature.

\section{Related works}

Examples of chaos-based tools for image recognition have been reported in the literature. For instance, in \cite{Y17} the authors approximate the digital image by a parametric Bezier function and carry out an analysis of the chaotic behavior resulting from repetitively iterating that function. In \cite{YW15}, a chaotic iteration procedure is proposed to reduce the impact of local dynamic changes and for an application to face recognition. Actually, these approaches share some similarities with the idea of chaotic maps.

Chaos theory has also been investigated for image recognition in association with other algorithms for pattern recognition, especially those based on evolutionary computing. Examples in this line are \cite{CI06} in particle swarm optimization and \cite{GWJWZ10} with genetic algorithms.

Another promising related approach in the last years has been cellular automata (CA). In \cite{LEA10}, the authors develop a direct CA modeling by binarizing the original image and iteratively applying well established transition functions over the CA lattice. More recently, an elaborated strategy has been proposed in \cite{SWBB15} where a physical model (corrosion) was used to formulate transition rules of a CA model. The method was applied to texture recognition achieving state-of-the-art results at the moment of the publication.

Bringing our attention specifically to chaos map modeling in computer vision, there is a variety of works on image encryption, steganography, watermar\-king and other security applications. However, for image analysis it is much more difficult to find relevant literature. In \cite{ZM01} we have an interesting application of a coupled network of chaotic maps for scene segmentation. The method relies on the principle that the time evolution of chaotic maps associated to the same object should be synchronized, as opposed to the desynchronized maps of different objects. For image recognition, we can mention \cite{YWA20}, where feature selection is carried out by chaotic whale optimization combined with chaotic maps. The method is applied to brain tumor classification. Logistic neural networks \cite{KO14} is a simplified model based on Hopfield networks, introducing a logistic map component to the neuronal activation, in this way gaining chaoticity (mathematically demonstrated) and interesting capabilities for pattern recognition and associative memory.   

\section{Background}

\subsection{Chaotic maps}\label{sec:maps}

Chaotic maps are the most canonical and easy to understand examples of chaotic systems. They are iterative maps (functions) that evolve in their independent variables exhibiting chaotic behavior, i.e., they are highly sensitive to initial conditions and seem to be irregular and random even though they are generated by a deterministic process. In this work we use six types of chaotic maps that are defined in the following.

The first one is the circle map. In its chaotic configuration, the evolution equation is given by
\begin{equation}
	x_{k+1} = x_k + \mu - (\nu/2\pi)\sin(2\pi x_k)\mathrm{mod}(1),
\end{equation}
where $\mathrm{mod}(1)$ in the remainder of an integer division by $1$. The combination $\mu=0.2$ and $\nu=0.5$ provides chaotic behavior and is used here.

The second one is Gauss map:
\begin{equation}
	x_{k+1} = \left\{
	\begin{array}{ll}
		0, & x_k=0\\
		1/x_k \mathrm{mod}(1), & \mbox{otherwise}.\\
	\end{array}
	\right.
\end{equation}

The third is probably the most well-known and studied chaotic map, i.e., the logistic map:
\begin{equation}
	x_{k+1} = \mu x_k(1-x_k).
\end{equation}
Several ranges for the real parameter $\mu$ can be used to generate chaotic behavior. Here we adopt $\mu=3.8$.

The fourth chaotic map employed in our experiments is the sine map, defined by
\begin{equation}
	x_{k+1} = \frac{\mu}{4}\sin(\pi x_k).
\end{equation}
Here we use $\mu=4$ for the objective of obtaining a chaotic evolution.

The fifth map is the Singer map, a representer of the important class of polynomial maps:
\begin{equation}
	x_{k+1} = \mu(7.86x_k-23.31x_k^2+28.75x_k^3-13.302875x_k^4).
\end{equation}
We adopt $\mu=1.07$ obtaining the desired chaotic evolution.

Finally, the sixth chaotic map used here is the Tent map:
\begin{equation}
	x_{k+1} = \left\{
	\begin{array}{ll}
	x_k/0.7, & x_k<0.7\\
	(10/3)(1-x_k), & \mbox{otherwise}.\\
	\end{array}
	\right.	
\end{equation}

\subsection{Local binary patterns}

Local binary patterns (LBP) \cite{OPM02} are local image descriptors that quantify how each pixel is related to its neighbors. It has been widely used especially in texture and face recognition, due to its simple implementation and high computational efficiency. The literature has presented a huge number of variations of the original method, each one with their own advantages and drawbacks. Here we employ one of the most widely used versions, which is the rotation-invariant and uniform method presented in \cite{OPM02}.

In this version, similar patterns are grouped together, reducing the number of features and also boosting the recognition performance, as demonstrated in \cite{OPM02}. The mathematical expression for a reference pixel with gray level $g_c$ can be summarized by
\begin{equation}
	LBP_{P,R}^{riu2} = 
	\left\{
		\begin{array}{ll}
			\sum_{p=0}^{P-1}H(g_p-g_c)2^p, & U(LBP_{P,R})\geq 2\\
			P+1, & \mbox{otherwise.}
		\end{array}
	\right.
\end{equation}
$R$ and $P$ are important parameters in the method and correspond, respectively, to the radius of a neighborhood whose center is the reference pixel and the number of pixels evaluated within that distance. $g_p$ are the gray levels of those $P$ pixels. As $P$ is frequently larger than the number of real pixels at distance $R$ from the center in the original image, more values of $g_p$ can be obtained by linear interpolation. $H(x)$ is the Heaviside step function: $H(x)=1$ if $x\geq 0$ and $H(x)=0$, otherwise. Finally, the rotation-invariant uniformity function is defined by
\begin{equation}
	U(LBP_{P,R}) = |H(g_{P-1}-g_c)-H(g_0-g_c)| + \sum_{p=1}^{P-1}|H(g_p-g_c)-H(g_{p-1}-g_c)|.
\end{equation}

\section{Proposed method}

This work proposes a new perspective for the representation of texture local descriptors. To exemplify the strategy we use LBP descriptors as they have simple implementation and interpretation and are very popular in texture recognition. Nevertheless, it is worth to highlight that the described approach can be easily adapted to any other local representation variations of LBP \cite{OPM02}, dense SIFT \cite{L04}, bag-of-visual-words \cite{VZ09}, an others.

The new perspective is accomplished by an appropriate application of chaotic maps. As described in Section \ref{sec:maps}, chaotic maps are functions chosen in such a way that their successive application converge to a chaotic behavior, i.e., with a slight variation in initial conditions implying large variation in the function output. Theoretically, our intention is to explore the chaotic behavior as a means to unveil complex non-linear patterns that are latent on the pixel distribution and that cannot be appropriately represented by the classical image descriptor. We could do that in two domains: the gray-level and the spatial arrangement of the pixel. In the first one, we would have a transform that would act in a unique manner over each pixel intensity, but the spatial correlation would be lost. But the local relation is known to be paramount for an effective image descriptor and LBP is a good example of that. On the other hand, a simple position rearrangement of pixels would obviously represent nothing more than a simple image processing operation, and wound not give any relevant contribution to the image descriptor.

Given that, we propose here a joint transform based on chaotic maps and combining spatial and gray-level domains. The general idea is quite straightforward and can be easily implemented. Inspired by geometrical approaches like fractal descriptors \cite{FCB18} and chaotic encryption \cite{CMC04}, we map the gray texture image onto 3D space, considering the gray value as the third coordinate of the pixel. To avoid any bias towards spatial or value domain, all values of spatial coordinates and pixel intensities are normalized in $[0,1]$. The ergodicity and Poincar\'{e} recurrence property often satisfied by these maps also make this matter less relevant in our context as at some iteration any point in the space-phase is supposed to be approximated as much as desired.

Given a grayscale normalized image $I_0 \in [0,1]^{m \times n} \rightarrow [0,1]$, we first define a matrix $x_0 \in \mathbb{R}^{mn \times 3}$ to represent the image pixels, such that for the $p^{th}$ pixel with coordinates $(i,j)$ \footnote{Here we adopt the convention of setting the first index of a vector/matrix as $1$, which is more consistent with mathematical notation and independent of programming language.} we have
\begin{equation}
	x_0(p,1) = i, \qquad x_0(p,2) = j, \qquad x_0(p,3) = I_0(i,j).
\end{equation}
In the following, we apply any of the maps $\mathfrak{M}:[0,1] \rightarrow [0,1]$ defined in Section \ref{sec:maps} (e.g., for the logistic map: $\mathfrak{M}(x) = \mu x(1-x)$). At each iteration the chaotic map $\mathfrak{M}$ is point-wise applied to the matrix $x$:
\begin{equation}
	x_{k+1} = \mathfrak{M}(x_k), \qquad k = 0,1,\cdots,n_{iter}-1,
\end{equation}
where $n_{iter}$ is the number of iterations. Here we fixed $n_{iter} = 10$.

Each matrix $x_k$ is therefore reconstructed into a transformed image $I_k$. This step needs to be carefully addressed as the map $\mathfrak{M}$ is not necessarily one-to-one. Besides, we should guarantee that the reconstructed matrix corresponds to a function (and consequently to an image), i.e., the same spatial coordinate cannot map to more than one value for the pixel intensity. For that, we first sort the unique values of $x_k(\cdot,1)$ and $x_k(\cdot,2)$, i.e., the spatial coordinates of all pixels at the $k^{th}$ iteration of the chaos map. Formally we define two sets of sorted unique values $\mathcal{U}_1$ and $\mathcal{U}_2$:
\begin{equation}
	\mathcal{U}_1(q) = r \mbox{ iff } \exists p : x_k(p,1) = r \mbox{ and } \mathcal{U}_1(q) < \mathcal{U}_1(q'), \forall q<q',
\end{equation}
\begin{equation}
	\mathcal{U}_2(q) = r \mbox{ iff } \exists p : x_k(p,2) = r \mbox{ and } \mathcal{U}_2(q) < \mathcal{U}_2(q'), \forall q<q',
\end{equation}
as well as the index sets $\mathcal{I}_1$ and $\mathcal{I}_2$:
\begin{equation}
	\mathcal{I}_1(q) = p \mbox{ iff } x_k(p,1) = \mathcal{U}_1(q),
\end{equation}
\begin{equation}
	\mathcal{I}_2(q) = p \mbox{ iff } x_k(p,2) = \mathcal{U}_2(q).
\end{equation}
The image $I_k$ is therefore constructed by
\begin{equation}
	I_k(q_1,q_2) = x_k(q,3),
\end{equation}
where $q_1$ and $q_2$ correspond to the indexes mapped for the new image dimensions:
\begin{equation}
	q_1 = \mbox{div}(\mathcal{I}_1(q),\mbox{dim}(\mathcal{I}_1))+1, \qquad q_2 = \mbox{mod}(\mathcal{I}_2(q),\mbox{dim}(\mathcal{I}_1))+1,
\end{equation}
being $\mathrm{div}(a,b)$ and $\mathrm{mod}(a,b)$ respectively the quotient and remainder of the integer division of $a$ by $b$ and $\mbox{dim}(\mathcal{I}_1)$ the number of components of the vector $\mathcal{I}_1$. 
%Formally it can be represented by
%\begin{equation}
%	\tilde{x_k}(\cdot,1) = \{\tilde{x}(p,1):\tilde{x}(p,1)\in x_k(\cdot,1)\}_{p=1,\cdots,mn},
%\end{equation}
%\begin{equation}
%	\tilde{x_k}(\cdot,2) = \{\tilde{x}(p,2):\tilde{x}(p,2)\in x_k(\cdot,1)\}_{p=1,\cdots,mn},
%\end{equation}
%where the set notation ensures uniqueness.

\begin{figure}[!htpb]
	\includegraphics[width=\textwidth]{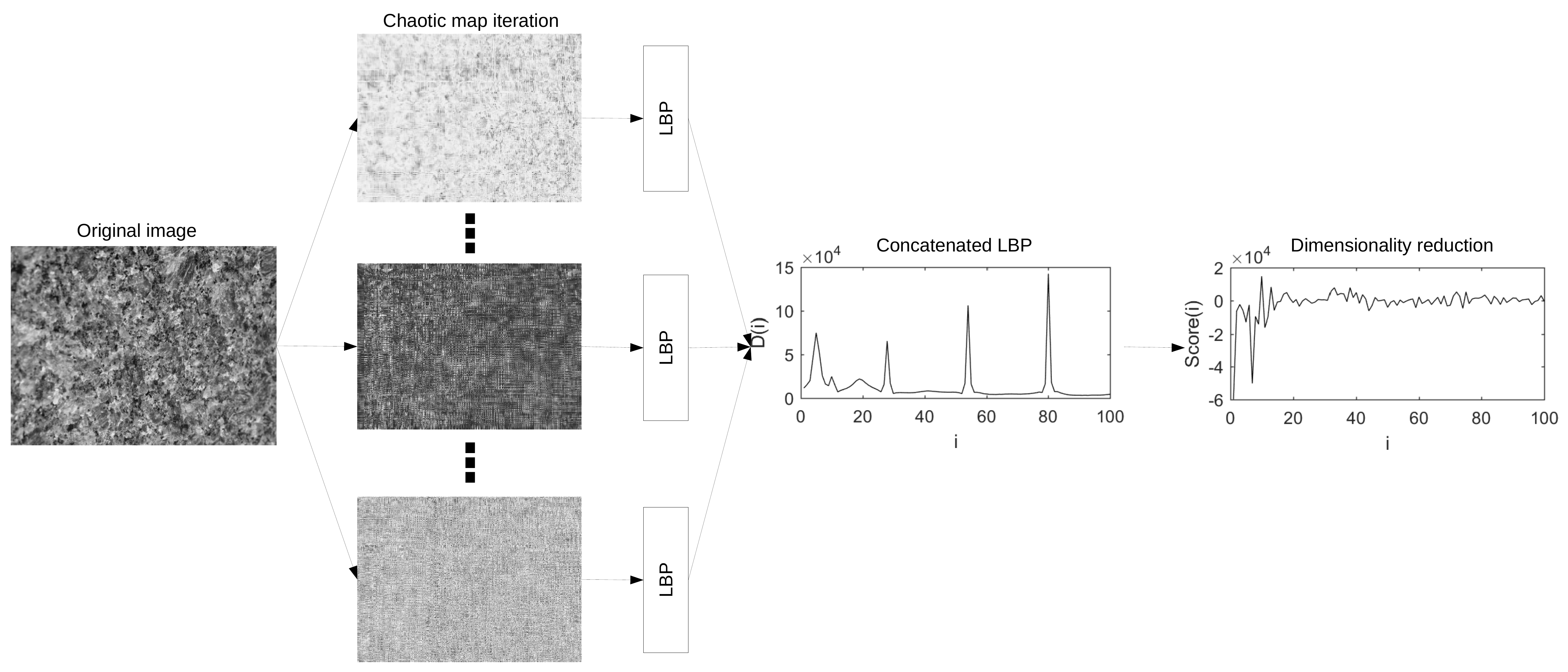}
	\caption{Proposed method.}
	\label{fig:method}
\end{figure}

For recognition purposes it is convenient to balance the influence of the maps, as their direct use would severely disorganize the image structure. Although this is a fundamental characteristic of chaos for cryptography, for recognition it would be disastrous. To accomplish such balanced chaos, we perform a weighted sum between the current image and the transformed one. In this way, the final descriptors are provided by the LBP distribution of the weighted images, i.e.:
\begin{equation}\label{eq:final_desc}
	\mathfrak{D}(I) = \bigcup_{\substack{k=1,\cdots,n_{iter}\\i=0,\cdots,1/\delta}} \mbox{LBP}((1-i\delta)I_{k-1} + i\delta I_k),
\end{equation}
where $\delta$ is a pre-defined parameter. Here we employ $\delta=0.1$ and such choice was empirically motivated. Using other values would be a mere trade-off. Larger values would rapidly disorganize the image structure, whereas smaller values could be used at the expense of more iterations necessary to capture the same degree of chaoticity. For the number of map iterations we adopt $n_{iter} = 10$. The pseudocode in Algorithm \ref{alg:chaosMap} (together with Table \ref{tab:routines} for the auxiliary routines) provides some more technical details concerning the implementation of the proposed method. Finally, to cope with the potentially large number of features collected by this procedure, classical techniques of dimension reduction, like principal component analysis \cite{B06}, can be used. Figure \ref{fig:method} visually illustrates the main steps involved in the methodology.

\begin{algorithm}[!htpb]
	\caption{Chaos map image descriptors.}
	\label{alg:chaosMap}
	\begin{algorithmic}[1] % The number tells where the line numbering should start	
		\Input $I$ (Gray-scale image); $map$ (Chaotic map, i.e., 'logistic', 'circle', etc.)
		\Output $D$ (Feature vector)
		\State $D \gets \emptyset$
		\State $n_{iter} \gets 10$
		\State $\delta \gets 0.1$	
		\For{$i = 1 \mbox{ \textbf{to} } \mathrm{size}(I,1)$}
			\For{$j = 1 \mbox{ \textbf{to} } \mathrm{size}(I,2)$}
				\State $p \gets (i-1)\mathrm{size}(I,2)+j$ \Comment{2D index to linear index}
				\State $x_0(p,1) \gets i/\mathrm{size}(I,1)$
				\State $x_0(p,2) \gets j/\mathrm{size}(I,2)$
				\State $x_0(p,1) \gets I(i,j)/255$ \Comment{assuming the maximum pixel value as 255}
			\EndFor
		\EndFor
		\For{$k = 1 \mbox{ \textbf{to} } n_{iter}-1$}
			\State $x_{k} \gets \mathrm{chaotic\_map}(x_{k-1},map)$
			\State $[\mathcal{U}_1,\mathcal{I}_1] \gets \mathrm{unique}(x_k(\cdot,1))$ 
			\State $[\mathcal{U}_2,\mathcal{I}_2] \gets \mathrm{unique}(x_k(\cdot,2))$
			\For{$q = 1 \mbox{ \textbf{to} } \mathrm{length}(\mathcal{I}_1)$}
				\State $q_1 \gets \mbox{div}(\mathcal{I}_1(q),\mbox{dim}(\mathcal{I}_1))+1$ 
				\State $q_2 \gets \mbox{mod}(\mathcal{I}_2(q),\mbox{dim}(\mathcal{I}_1))+1$
				\State $I_k(q_1,q_2) \gets x_k(q,3)$
			\EndFor
			\For{$r = 0 \mbox{ \textbf{to} } 1/\delta$}
				\State $D \gets D \cup \mathrm{LBP}((1-i\delta)I_{k-1} + i\delta I_k)$
			\EndFor			
		\EndFor		
	\end{algorithmic}
\end{algorithm}

\begin{table}[!htpb]	
	\centering
	\caption{Auxiliary functions and notation used by Algorithm \ref{alg:chaosMap}.}	
	\begin{tabular}{ll}
		\hline
		$\mbox{size}(x,n)$ & Size of the $n^{th}$ dimension of the array $x$.\\
		$[u,i] = \mbox{unique}(x)$ & Unique elements of $x$ returned in $u$ and the respective indexes returned by $i$.\\
		\hline
	\end{tabular}
	\label{tab:routines}	
\end{table}

\section{Theoretical analysis}

A closed formula would certainly be the easiest way to interpret the action of chaos maps over real-world data. Unfortunately, it is well known that in general no such formula exists for any chaotic map, which to a great extent is consequence of the unpredictable nature of chaos. There are, however, a few exceptions and logistic map is an example in the particular cases of $\mu=-2$, $\mu=2$, and $\mu=4$. In all these cases a closed form to calculate $x_n$ follows a general expression:
\begin{equation}\label{eq:general_logistic}
	x_n = \frac{1}{2}\left\{ 1-f_\mu \left[ \mu^n f_\mu^{-1}(1-2x_0) \right] \right\},
\end{equation}
where the function $f(x)$ has been exactly found in the literature \cite{M20} for the cases in Table \ref{tab:fx}.
\begin{table}[!htpb]
\centering
\caption{$f(x)$ of logistic map in the general expression (\ref{eq:general_logistic}) for some values of $\mu$.}
\label{tab:fx}
\begin{tabular}{|c|c|}
	\hline
	$\mu$ & $f(x)$\\
	\hline
	$-2$ & $2\cos\left( \frac{1}{3}\left( \pi-\sqrt{3}x \right) \right)$\\
	$2$ & $\mathrm{e}^x$\\
	$4$ & $\cos x$\\
	\hline
\end{tabular}
\end{table}

It has also been conjectured that such general expression could be valid for other (possibly all) values of $\mu$. This has been explored in \cite{M20}, where the authors take advantage of (\ref{eq:general_logistic}) to find a representative power series
\begin{equation}\label{eq:power} 
	F_{\mu}(x) = \sum_{n=0}^{\infty}a_nx^n.
\end{equation}
Despite accepting that such power series do not correspond to any elementary function as in Table \ref{tab:fx}, the polynomial expansion still is a worthwhile tool to gain some intuition on the map dynamics.

To obtain the power series, a more suitable form of (\ref{eq:general_logistic}) would be
\begin{equation}\label{eq:xnF}
	x_n = F_{\mu}\left( (\sqrt{\mu})^n F_{\mu}^{-1}(x_0) \right),
\end{equation}
subject to the constraint $F_{\mu}(0) = 0$. As described in \cite{M20}, $a_0$ should be $0$ to satisfy the constraint, all the odd coefficients in (\ref{eq:power}) are zero, $a_2$ is arbitrary and can be set to $1$ for simplicity, and the remaining coefficients are obtained by the recursion
\begin{equation}
	a_{2n} = \frac{1}{\mu^{n-1}-1}\sum_{j=1}^{n-1}a_{2j}a_{2n-2j}, \qquad n = 2,3,\cdots.
\end{equation}
Working with $\mu>1$ as we do here, we ensure that (\ref{eq:power}) is an alternating series with guaranteed convergence. A second order approximation would be in this case:
\begin{equation}\label{eq:Fmu}
	F_{\mu}(x) \approx x^2 - \frac{1}{\mu-1}x^4,% + \frac{2}{(\mu-1)^2(\mu+1)}x^6 - \frac{5+\lambda}{(\mu-1)^3(1+\mu)(1+\mu+\mu^2)}x^8,
\end{equation}
with an upper bound estimate for the error $E$ of an alternating series given by
\begin{equation}
	%E_8(x) \leq \left| \frac{2(7+3\mu+2\mu^2)}{(-1+\mu)^4(1+\mu)^2(1+\mu^2)(1+\mu+\mu^2)}x^{10} \right|,
	E_4(x) \leq \left| \frac{2}{(\mu-1)^2(\mu+1)}x^{6} \right|,
\end{equation}
which for $\mu=3.8$ and $x\in[0,1]$ would result in
\begin{equation}
	E_8(x) \leq 0.05,
\end{equation}
a reasonable approximation for our purposes.

For a complete expression of (\ref{eq:xnF}) we still need $F_\mu^{-1}(x)$. 
%We could find a cumbersome expression using (\ref{eq:Fmu}), but we prefer a more general approach by  
%
%It is also of our interest to obtain an inverse function for $x_n$, i.e., some expression that provides $x_0$ given $x_n$. For that, we apply the theorem of inverse of composite functions \cite{}, to obtain
%\begin{equation}
%	x_0 = F_{\mu}\left( \frac{1}{\sqrt{\mu}^n} F_{\mu}^{-1}(x_n) \right),
%\end{equation}
Considering the $2^{nd}$ order approximation, it can be easily solved explicitly as a biquadratic equation:
\begin{equation}
	F = z - \alpha z^2,
\end{equation}
where $z = x^2$ and $\alpha = \frac{1}{\mu-1}$. The root is
\begin{equation}
	z = \frac{1\pm\sqrt{1-4\alpha F}}{2\alpha}
\end{equation}
and
\begin{equation}
	x = \sqrt{\frac{1\pm\sqrt{1-4\alpha F}}{2\alpha}}
\end{equation}
Plugging into (\ref{eq:xnF}) and (\ref{eq:Fmu}) we have
\begin{equation}
	\begin{array}{l}
	x_n = F \left( \mu^{n/2}\sqrt{\frac{1\pm\sqrt{1-\frac{4x_0}{\mu-1}}}{\frac{2}{\mu-1}}} \right) = \\ = \mu^n \frac{1\pm\sqrt{1-\frac{4x_0}{\mu-1}}}{\frac{2}{\mu-1}} - \frac{1}{\mu-1}\mu^{2n}\left( \frac{1\pm2\sqrt{1-\frac{4x_0}{\mu-1}} + 1-\frac{4x_0}{\mu-1}}{\frac{4}{(\mu-1)^2}} \right)
	\end{array}
\end{equation}

Rearranging and simplifying:
\begin{equation}\label{eq:closed_approx}
	\begin{array}{l}	
	x_n = \left( \frac{\mu^{n+1}-\mu^n}{2} \right) \left( 1 \pm\sqrt{1-\frac{4x_0}{\mu-1}} \right) - \\ 
	\left( \frac{\mu^{2n+1} - \mu^{2n}}{2} \right) \left( 1\pm\sqrt{1-\frac{4x_0}{\mu-1}}-\frac{2x_0}{\mu-1} \right).
	\end{array}
\end{equation}
In terms of practical computation, a variety of numerical techniques could be employed to address convergence issues in this expression \cite{M20}. Nevertheless, here we are more interested on the asymptotic behavior of (\ref{eq:closed_approx}). First it is interesting to observe the role of $\frac{x_0}{\mu-1}$ on the iteration. This is a fixed value and the largest it is (closest to $1$), the more rapidly $x_n$ will increase. Another important factor is $\mu$. Again, if $\mu>1$, the largest its value the steepest is $x_n$ path. $\mu$ is actually significantly more influential than $x_0$ as it works as a multiplier raised to the $n^{th}$ power. This is also connected with the observation that chaotic behavior arises for values of $\mu>3$, corresponding to quicker growing of $x_n$ iteration.   

With regards to $x_0$, which is the most important parameter for our purposes, its influence is restrict to the expression
\begin{equation}
	\left( 1\pm\sqrt{1-\frac{4x_0}{\mu-1}} \right)^k, \qquad k = 1,2,\cdots.
\end{equation}
If this is taken separately, the summation over $k$ presents quasi-linear asymptotic behavior. This is an interesting finding as it implies that the value of $x_0$ does not contribute significantly to the chaotic evolution. The non-linearity is essentially guaranteed by the action of $\mu$. This is quite convenient for a texture local descriptor as it ensures that the chaotic dynamics is relatively agnostic to pixel arrangements and image regions. Together with the exponential dependence on $\mu$, the power series representation highlights one of the most important characteristics of a chaotic map for our purposes, which is the ability of making subtle variations more evident, either in spatial or intensity domain. The richness of the local representation is increased in this way and more complete and robust descriptors are a natural consequence of that transformation.

\section{Experiments}\label{sec:exp}

The performance of the proposed methodology on texture classification has been assessed on four databases.

KTHTIPS-2b \cite{HCFE04} is a set of color images representing 11 types of materials (classes), each class containing 432 images. Here the images are converted to gray scales. Each image has size $200 \times 200$. Two main points make this database particularly challenging. The first one is the focus on the material depicted in the image rather than on the visual appearance. This causes the same class to have images with substantially different patterns. The second is the training/testing protocol usually found in the recent literature for that database. Each class is evenly divided into 4 samples, and we always use 1 sample for training and 3 samples for testing. Each sample contains particular settings for illumination, scale and viewpoint. As a consequence, the algorithm is required to predict the class of images with patterns that are significantly different from those learned on the training set. 

UIUC \cite{LSP05} comprises a set of grayscale textures collected under non-controlled conditions of illumination, scale and pose. It is composed by 25 classes, each one with 40 images, each image with a resolution $640 \times 480$. The training/testing split consists of randomly dividing 50\% of samples for training and the remaining 50\% for testing. This is repeated 10 timed to compute average accuracy and standard deviation. 

UMD \cite{XJF09} is very similar to UIUC, except for the slightly more complex patterns of textures and the higher resolution ($1280 \times 960$). The number of images and classes as well as the standard training/testing protocol are exactly the same of UIUC.

\begin{figure}[!htpb]
	\includegraphics[width=\textwidth]{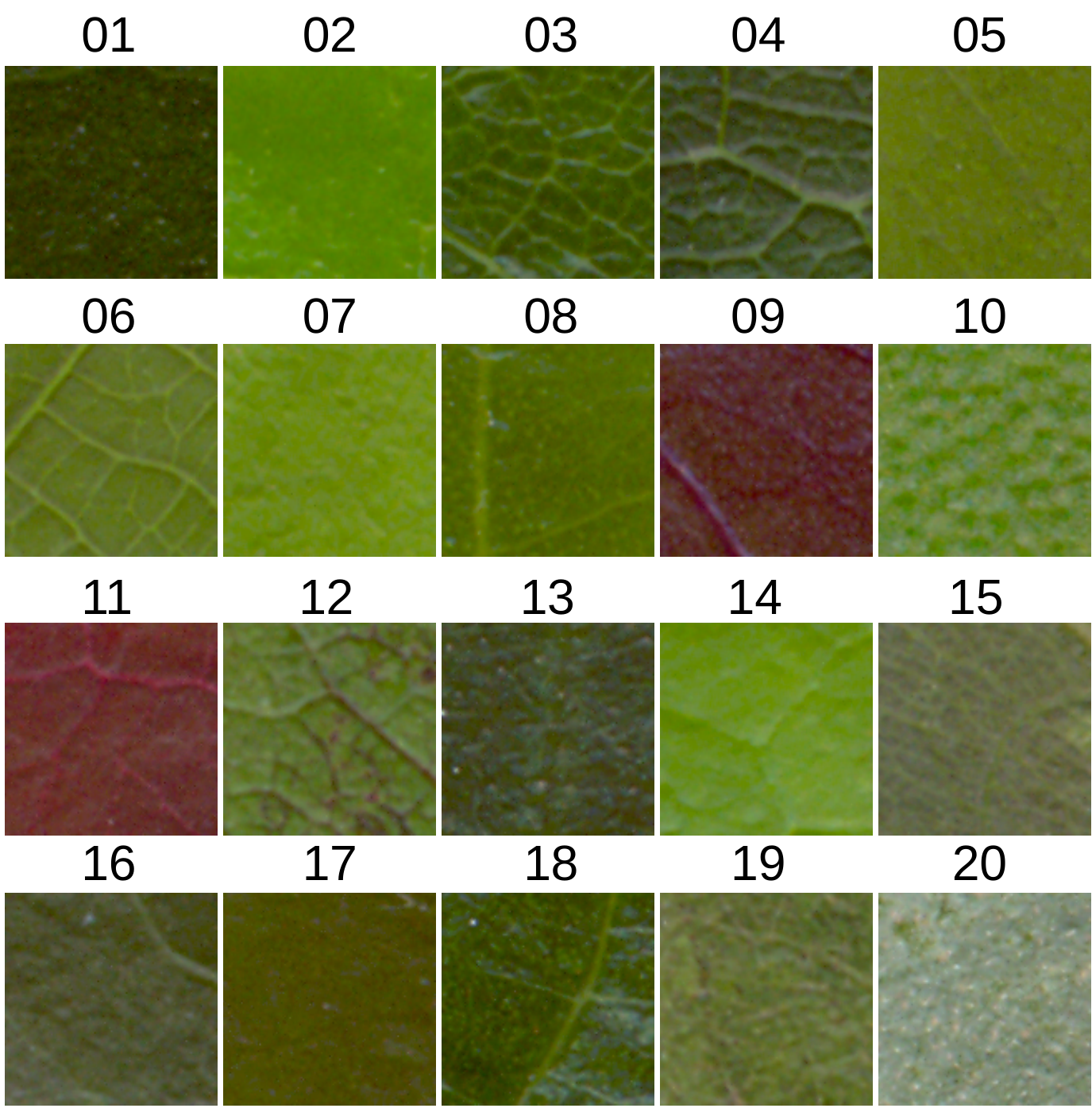}
	\caption{1200Tex texture samples, one for each class. Here we use a grayscale version of the database.}
	\label{fig:tex1200}
\end{figure}

1200Tex \cite{CMB09} is a collection of images of foliar surfaces of 60 Brazilian plant species. Each species correspond to a class in the data set and we have 60 images per class. These images correspond to 3 non-overlapping windows collected from each individual specimen, appropriately cleaned, photographed by a commercial scanner under rigorous conditions of illumination and scale and post-processed. Figure \ref{fig:tex1200} illustrates one sample image for each class. The trai\-ning/testing division is the same one adopted for UIUC and UMD, i.e., 50\% for training and 50\% for testing.

Finally, for the last step in the recognition system, i.e., the classifier, we verified the performance of three popular solutions: random forests (RF) \cite{H95}, support vector machine (SVM) with linear kernel \cite{CV95} and linear discriminant analysis (LDA) \cite{M04}. The optimal hyper-parameters were determined by cross-validation over the training set.

\section{Results and Discussion}

Given that our algorithm can run on intermediate steps between successive iterations, we define here a parameter $\alpha$, given by $k+i\delta$ in the notation of (\ref{eq:final_desc}). In this way, if $\alpha$ is an integer number, it corresponds to the number of iterations, otherwise, it represents an intermediate step between two iterations.

Table \ref{tab:classif} lists the classification accuracies for the three compared classifiers: RF, SVM and LDA. In this test we use logistic map and $\alpha=2.0$. Discriminant analysis consistently provided the highest accuracy in the analyzed databases. RF was the second best alternative. Here there are basically two reasons for the effectiveness of LDA. The first one is the large dimension of LBP features used to compose the feature vector. At the same time, such descriptors are not necessarily redundant and may still collaborate with the final classification. In this way, the projection taking into account the inter and intra-class variances is more suitable. Based on this, we proceed with the remaining tests using LDA classifier. 
\begin{table}[!htpb]
	\centering
	\caption{Accuracy (\%) for different classifiers in the benchmark databases (logistic map with $\alpha=0.2$).}
	\label{tab:classif}
	\begin{tabular}{cccc}
		\hline
		Database & RF & SVM & LDA\\
		\hline
%		KTH-TIPS2b & 58.9 & 61.5 & 63.6\\
		KTH-TIPS2b & 61.7 & 58.4 & \textbf{65.9}\\
		UIUC & 94.6 & 94.2 & \textbf{96.3}\\
		UMD & 98.0 & 97.4 & \textbf{99.5}\\
		1200Tex & 82.7 & 81.4 & \textbf{87.4}\\
		\hline
	\end{tabular}
\end{table}

\begin{figure}[!htpb]
	\centering
		\begin{tabular}{cc}
			KTHTIPS-2b & UIUC\\
			\includegraphics[width=.48\textwidth]{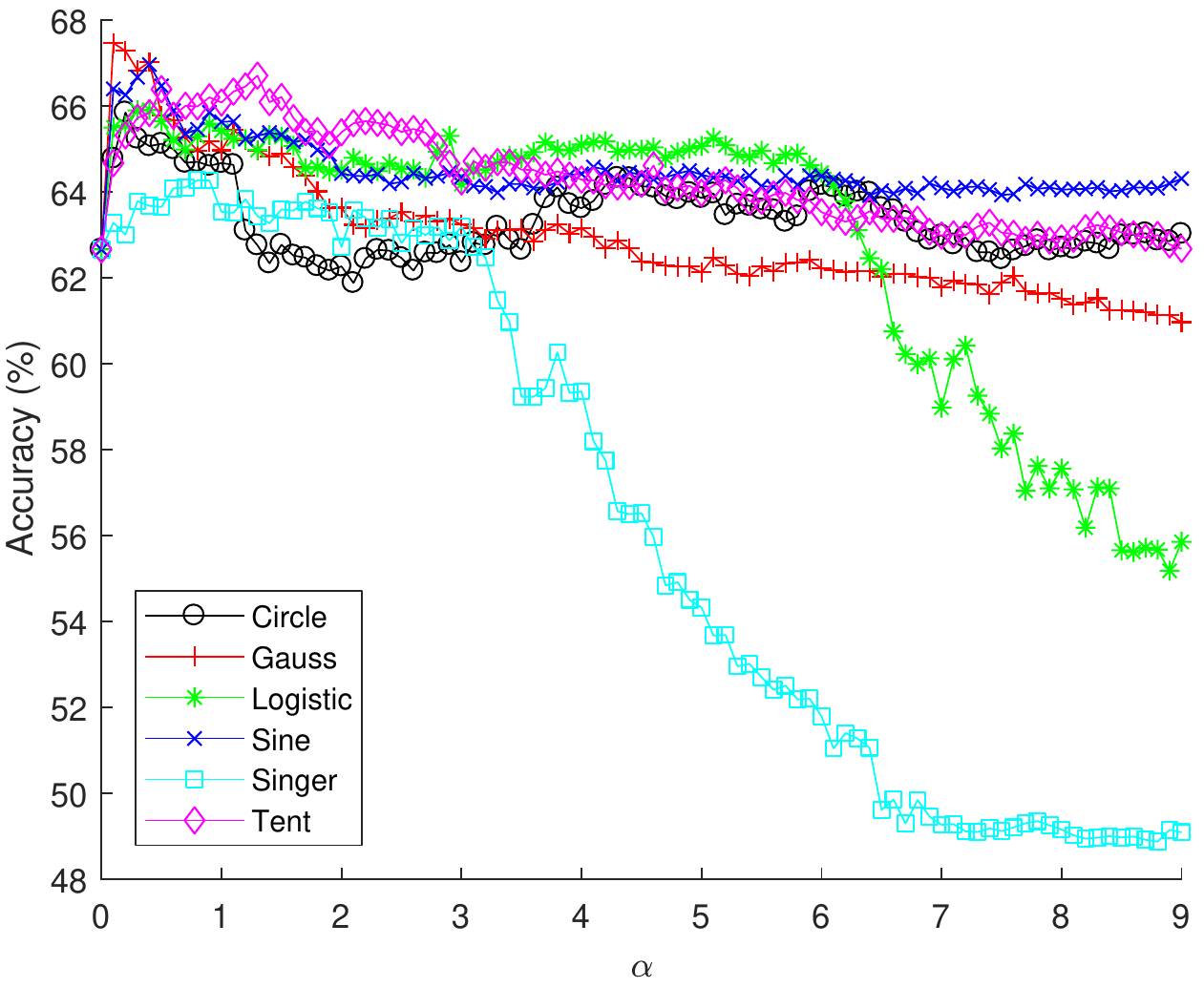}&
			\includegraphics[width=.48\textwidth]{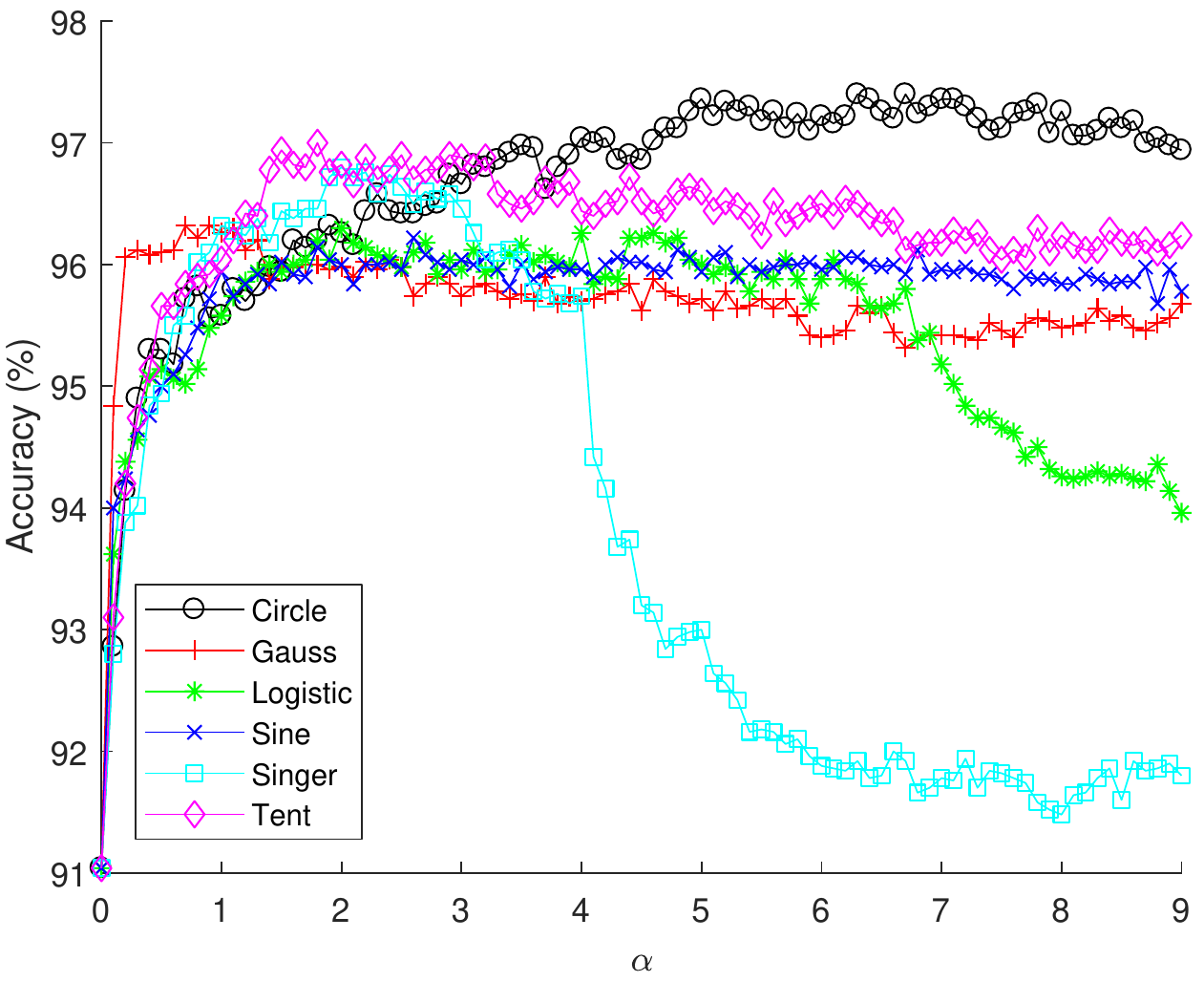}\\		
			UMD & 1200Tex\\	
			\includegraphics[width=.48\textwidth]{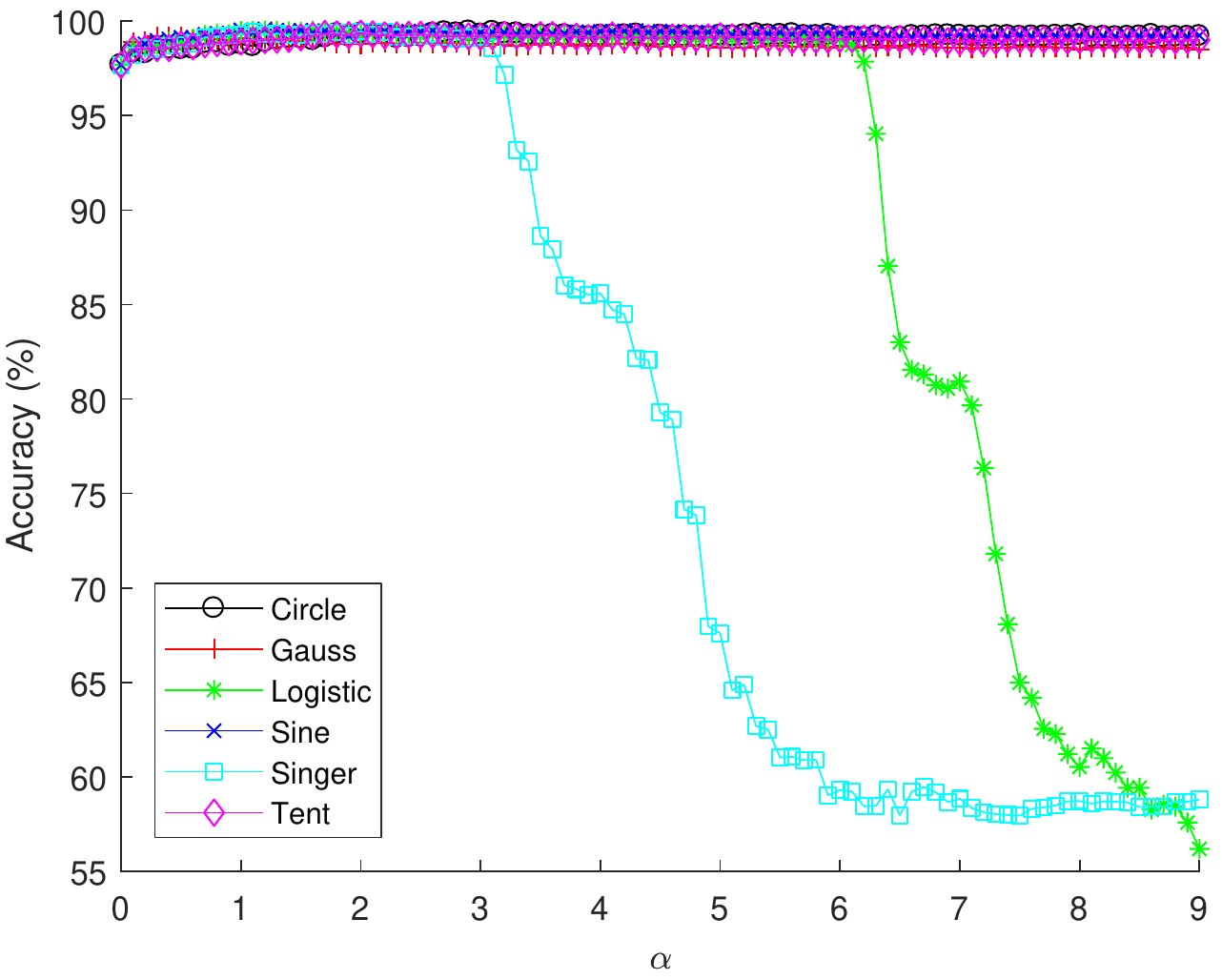}&
			\includegraphics[width=.48\textwidth]{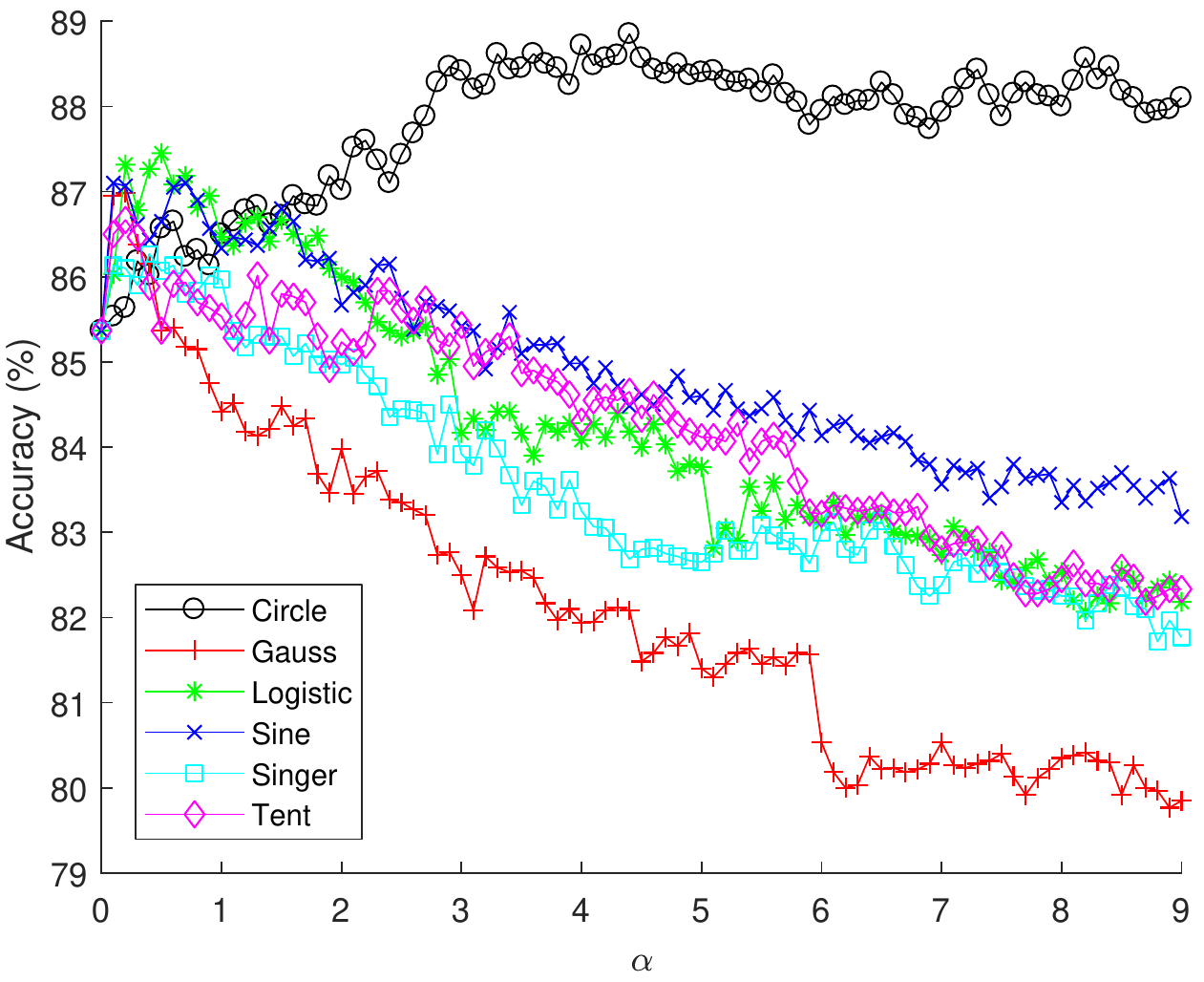}\\			
		\end{tabular}
	\caption{Accuracy in the analyzed databases using different chaotic maps and $\alpha$ values.}
	\label{fig:SR_all_maps}
\end{figure}

Figures \ref{fig:SR_all_maps} shows the accuracies for the different chaos maps and $\alpha$ values considered here. For the most challenging data sets, i.e., KTHTIPS-2b and 1200Tex we also used downsampled versions of each texture. More precisely, we combined features of the original image with features of the same image with dimensions reduced by a factor of $0.75$ and $0.50$. The different maps exhibit similar behavior in most data sets, with logistic and Singer maps having their performance rapidly degrading as the number of iterations (and $\alpha$ consequently) increases. This can be explained by the growing rate of the map. Logistic is a quadratic map, whereas Singer follows a quartic function. An expected consequence in these cases is more severe rearrangements of pixel patterns, surpassing the amount necessary for the comprehension of non-linear patterns. Chaos maps on 1200Tex, on the other hand, presented particular characteristics. Only circle map showed consistency in accuracy here. This is mostly motivated by the high homogeneity of the textures. Circle map, in this context, has strong periodicity and small increments at each time step, due to the small value 0.5 for the parameter $\nu$. Such smooth transform over the original image is adequate to obtain the controlled chaos pursued here.

Table \ref{tab:classif} lists the highest accuracies obtained by each chaotic map. Interestingly, for each database a different map was the best choice in terms of classification precision. Circle map, however, was a suitable option in most cases. As previously mentioned, periodic sinusoid maps are easier to control than higher-order polynomial maps. Such control is the basic principle behind the mechanism of non-linear representation explored here.
\begin{table}[!htpb]
	\centering
	\caption{Accuracy (\%) for different chaotic maps in the benchmark databases using the best possible $\alpha$ value.}
	\label{tab:classif}
	\begin{tabular}{ccccccc}
		\hline
		Database & Circle & Gauss & Logistic & Sine & Singer & Tent\\
		\hline
		% KTH-TIPS2b (no multiscale) & 62.0 & 63.9 & 62.4 & 61.6 & 60.5 & 61.6\\
		KTH-TIPS2b & 65.8 & \textbf{67.5} & 65.9 & 67.0 & 64.3 & 66.7\\
		UIUC & \textbf{97.4} & 96.3 & 96.3 & 96.2 & 96.8 & 97.0\\
		UMD & 99.5 & 99.2 & 99.5 & \textbf{99.6} & 99.5 & 99.3\\
		1200Tex & \textbf{88.8} & 87.0 & 87.4 & 87.1 & 86.3 & 86.7\\
		\hline
	\end{tabular}
\end{table}

\begin{figure}[!htpb]
	\centering
	\begin{tabular}{cc}
		\includegraphics[width=.48\textwidth]{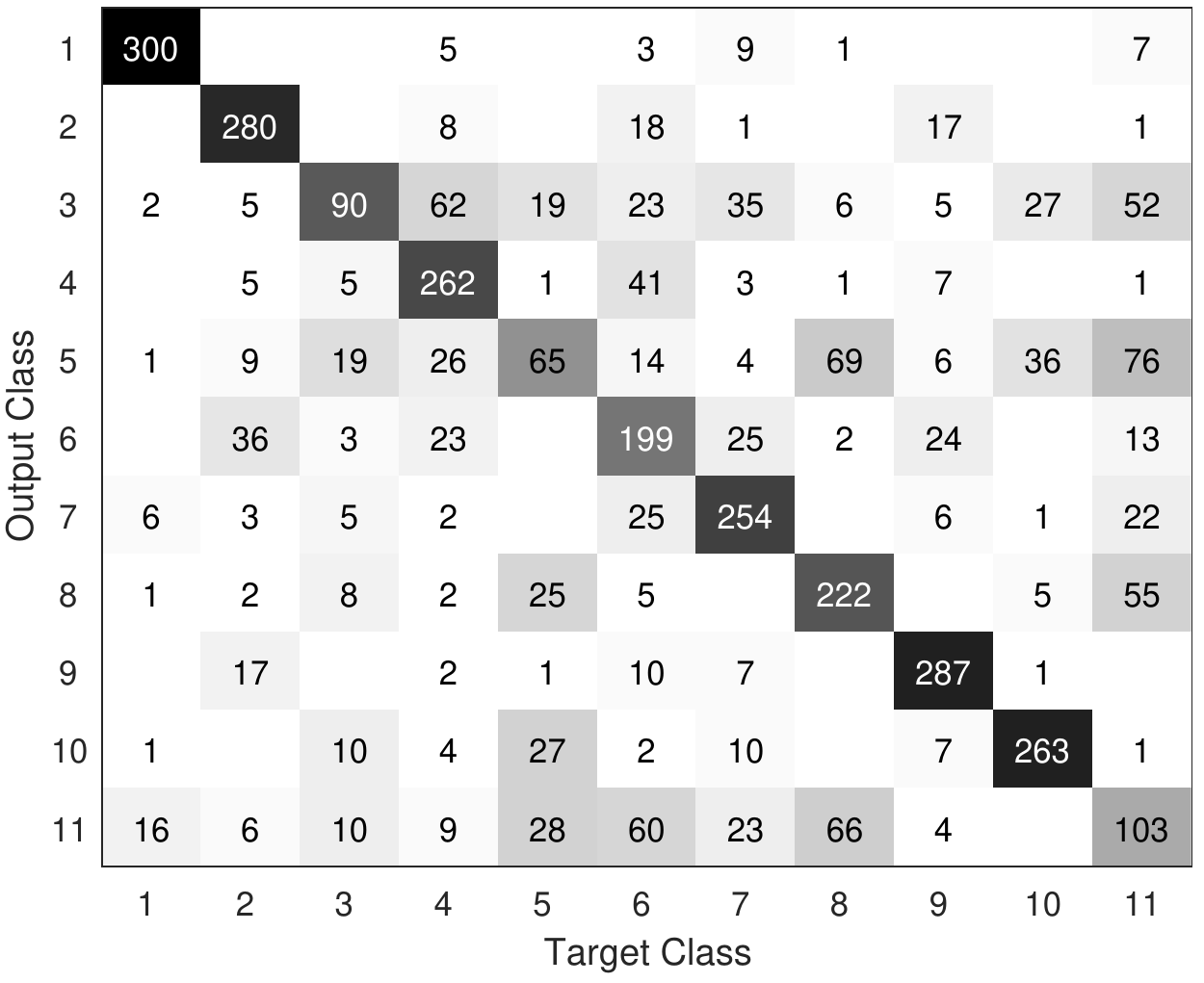} &
		\includegraphics[width=.48\textwidth]{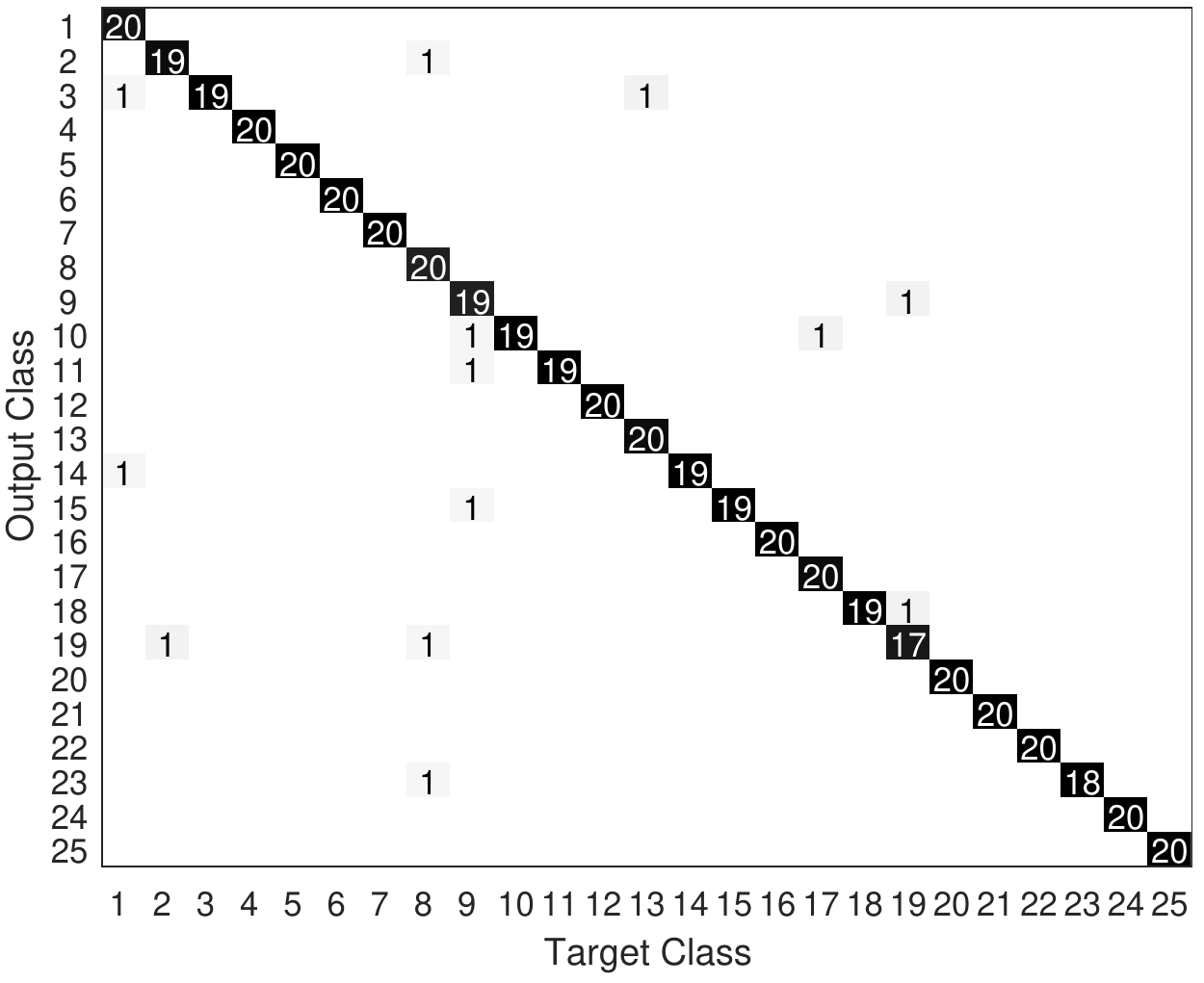}\\
		KTHTIPS-2b & UIUC\\
		\includegraphics[width=.48\textwidth]{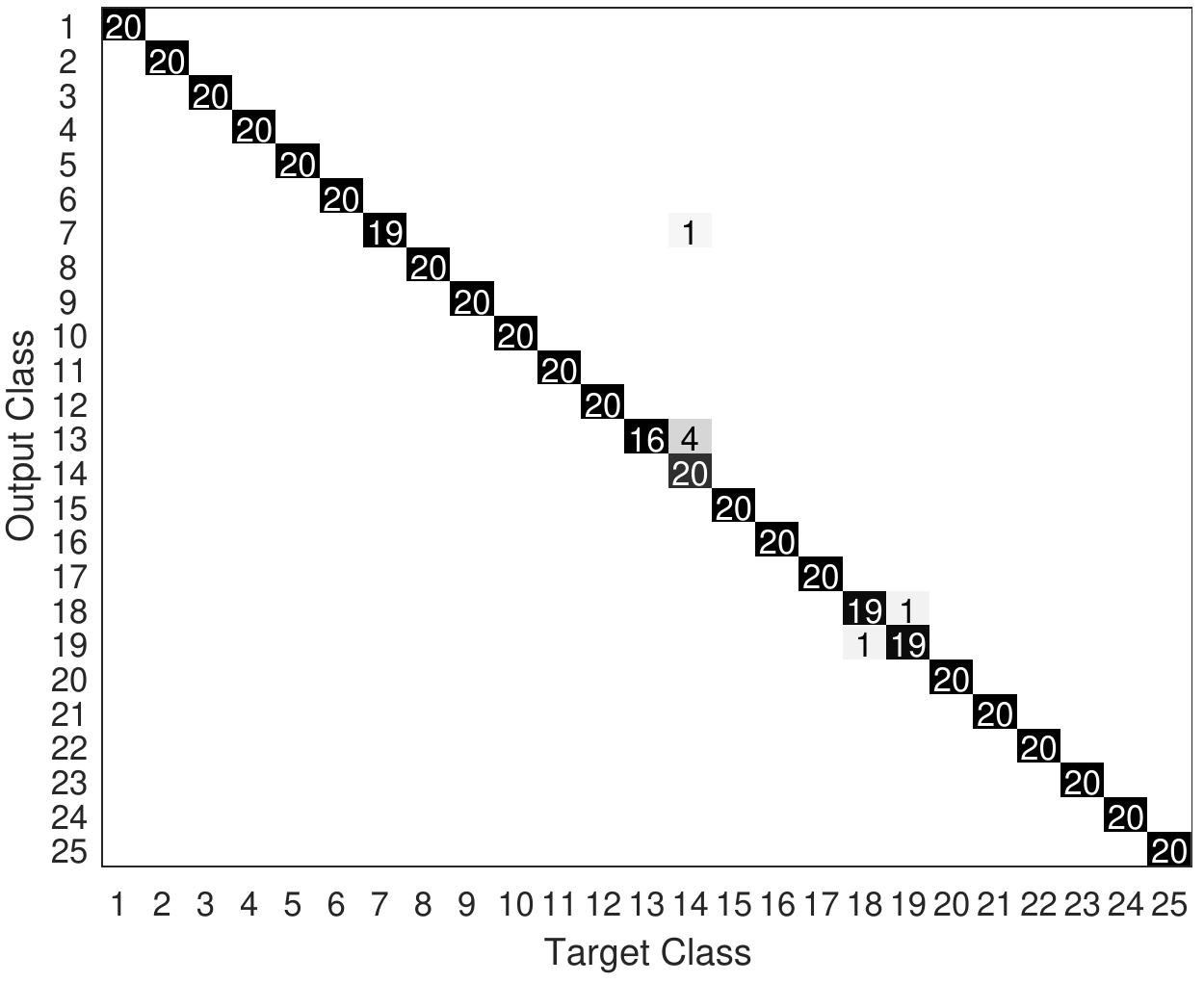} &
		\includegraphics[width=.48\textwidth]{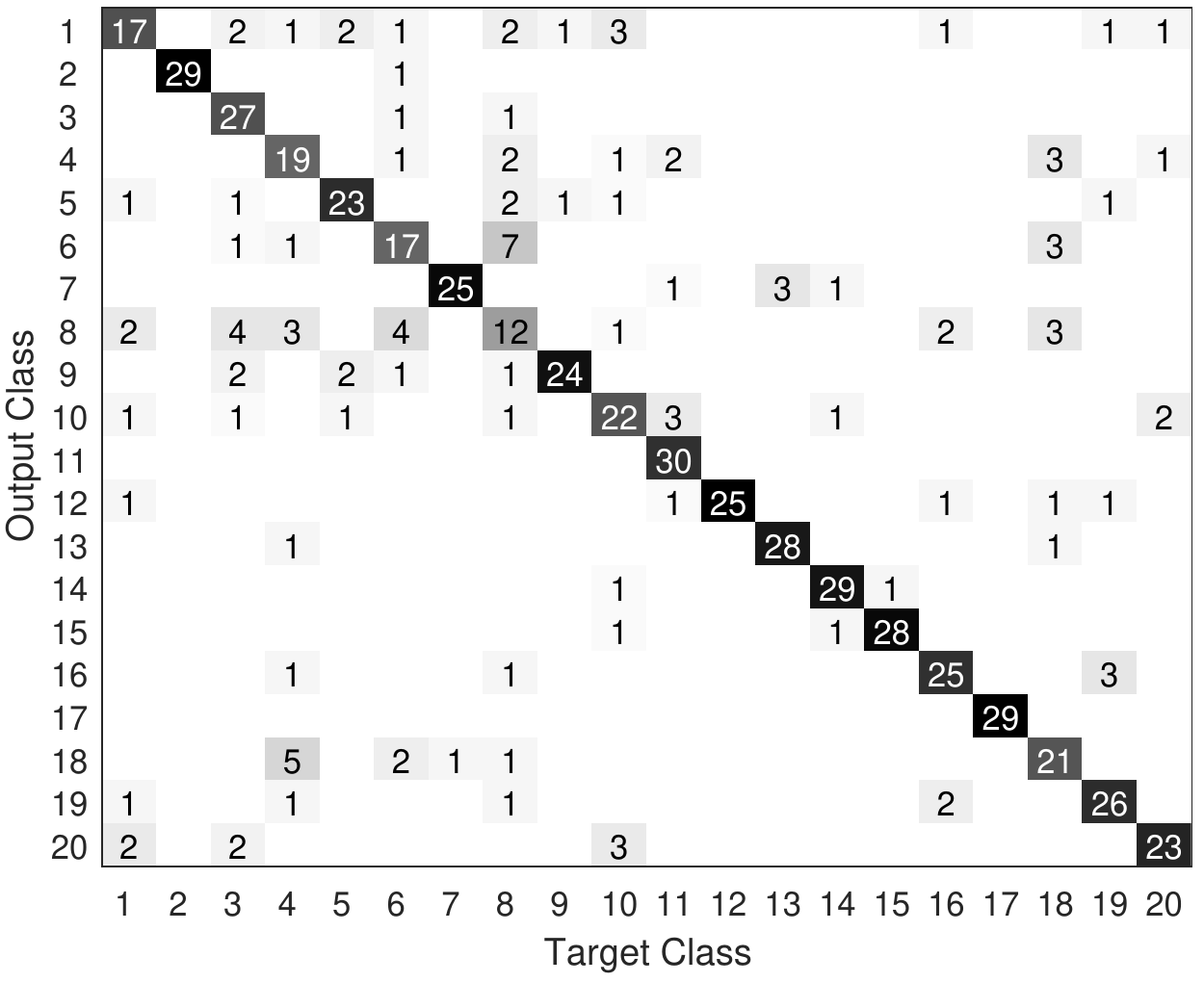}\\		
		UMD & 1200Tex\\
	\end{tabular}
	\caption{Confusion matrices for the proposed method on the texture databases.}
	\label{fig:CM}
\end{figure}

Figure \ref{fig:CM} depicts the confusion matrices of the proposed method in the compared texture collections. By showing the expected (target) class and the output of the classifier, we can have a good idea of the performance of the algorithm on different classes and, possibly, obtaining better understanding on the way that the method actually works. There is not too much to discuss here on UIUC and UMD as they are pretty close to the perfect classification and the misclassified samples change in each validation round, making them not relevant in statistical terms. On the other hands, KTHTIPS-2b and 1200Tex are substantially more challenging, as already noticed before, and deserve special attention. For the first one, we notice dramatic difficulties in classes 3, 5, and 11. The respective materials are ``corduroy'', ``cotton'' and ``wool''. These are all types of fabric and the confusion among these groups is expected. On 1200Tex, the confusion between classes 6 and 8 is noticeable. As illustrated in Figure \ref{fig:tex1200}, both classes share similar green tonalities. This is even more challenging here as we work on gray scales. Furthermore, those plant species also present similar distribution of nervures over the foliar surface. This is known to be a very important trait for the biological distinction among species. 

Tables \ref{tab:SR_SoA} and \ref{tab:SR_SoA2} exhibit the accuracies compared to other methods for texture classification published in the literature. We opted for separating results of 1200Tex as these are significantly less explored in the literature, given its more specific scientific domain. The proposed method confirms its potential, outperforming the compared methods, including those based on state-of-the-art convolutional neural networks (CNN) in UIUC and UMD. It is worth to emphasize here that UIUC and UMD are typical example of what can be considered as textures in its more strict sense, such that this great performance is a remarkable achievement.

As for KTHTIPS-2b, this is naturally a more complicated scenario for the proposed approach given its focus on material instances, as explained in Section \ref{sec:exp}. Here there are two main points that substantially favor learning-based approaches like FC-CNN. First the high dependence on color information, that is not considered in the present proposal. The second one is the possibility of pre-training on ImageNet, which despite all computational burden required, exposes the algorithm to a myriad of objects from the most diverse domains and makes it much more familiar with KTH materials. Nevertheless, even in this scenario the achieved performance can be considered competitive. We should also observe that an extension of the current method to address color information and combination with CNN to take advantage of transfer learning are interesting possibilities to be explored in future works. For now, the focus is on verifying and comprehending the action of chaos maps on texture images. 

A remarkable result here is the performance on the application of identifying plant species (Table \ref{tab:SR_SoA2}). Here we see our proposal outperforming sophisticated CNN-based approaches, like FV-CNN VGGVD, which frequently achieves the highest reported accuracy in several texture databases, but at the price of high computational burden and nearly impossible interpretation of the model. Other computationally intensive solutions like SIFT$+$VLAD are also outperformed.
\begin{table}[!htpb]
	\centering
	\caption{State-of-the-art accuracies on the benchmark data sets.}
	\label{tab:SR_SoA}
	\begin{tabular}{lccc}
			\hline
			Method & KTH2b & UIUC & UMD\\
			\hline
			MFS \cite{XJF09} & - & 92.7 & 93.9\\			
			VZ-MR8 \cite{VZ05} & 46.3 & 92.9 & - \\
			LBP \cite{OPM02} & 50.5 & 88.4 & 96.1\\
			VZ-Joint \cite{VZ09} & 53.3 & 78.4 & - \\
			BSIF \cite{KR12} & 54.3 & 73.3 & 96.1\\
			CLBP \cite{GZZ10} & 57.3 & 95.7 & 98.6 \\
			L. Liu et. al – CS \cite{LFK11} & - & 96.3 & 99.1\\
			PLS \cite{QXSL14} & - & 96.6 & 99.0\\
			SIFT+LLC \cite{CMKMV14} & 57.6 & 96.3 & 98.4\\					
			SIFT + KCB \cite{CMKMV14} & 58.3 & 91.4 & 98.0\\
			SIFT + BoVW \cite{CMKMV14} & 58.4 & 96.1 & 98.1\\									
			LBP$_{riu2}$/VAR \cite{OPM02} & 58.5 & 84.4 & 95.9\\
			PCANet (NNC) \cite{CJGLZM15} & 59.4 & 57.7 & 90.5\\
			RandNet (NNC) \cite{CJGLZM15} & 60.7 & 56.6 & 90.9\\
			SIFT + VLAD \cite{CMKMV14} & 63.1 & 96.5 & 99.3\\
			ScatNet (NNC) \cite{BM13} & 63.7 & 88.6 & 93.4\\
			OBIFs INNC \cite{TG12} & 66.3 & - & -\\			
			SIFT+IFV \cite{CMKMV14} & 69.3 & 97.0 & 99.2 \\	
			Xu et al.-OTF \cite{XHJF09} & - & 97.4 & 98.5\\	
			DeCAF \cite{CMKMV14} & 70.7 & 94.2 & 96.4\\
			SIFT + FV \cite{CMKV16} & 70.8 & - & -\\
			FC-CNN VGGM \cite{CMKV16} & 71.0 & 94.5 & 97.2\\			
			FC-CNN AlexNet \cite{CMKV16} & 71.5 & 91.1 & 95.9\\			
			(H+L)(S+R) \cite{LSP05} & - & 97.0 & 96.9\\			
			\hline
			Proposed & 67.5 & 97.4 & 99.6\\
			\hline
	\end{tabular}
\end{table}

\begin{table}[!htpb]
	\centering
	\caption{State-of-the-art accuracies for 1200Tex.}	
	\label{tab:SR_SoA2}
	\begin{tabular}{lc}
		\hline
		Method & Accuracy (\%)\\
		\hline
		LBPV \cite{GZZ10b} & 70.8\\
		Network diffusion \cite{GSFB16} & 75.8\\
		FC-CNN VGGM \cite{CMKV16} & 78.0\\		
		FV-CNN VGGM \cite{CMKV16} & 83.1\\		
		Gabor \cite{CMB09} & 84.0\\
		FC-CNN VGGVD \cite{CMKV16} & 84.2\\
		Schroedinger \cite{FB17} & 85.3\\		
		SIFT + BoVW \cite{CMKMV14} & 86.0\\		
		FV-CNN VGGVD \cite{CMKV16} & 87.1\\	
		SIFT+VLAD \cite{CMKMV14} & 88.3\\													
		\hline
		Proposed & 88.8\\
		\hline
	\end{tabular}
\end{table}	

In general, chaotic maps demonstrated to be useful for texture recognition, despite its well-established association with encryption and disarrangement of patterns rather than with pattern recognition. The success of the presented proposal is due mainly to the strict control of the chaos action by using a weight parameter balancing the effect of the chaotic map in one iteration with the outcome of the previous iteration. This is the idea that we call ``controlled chaos''. Such mechanism allows for the highlighting of nonlinear relations that could not be captured by the direct local descriptor at the same time that it preserves basic structures of the image that are necessary for an efficient description of the local image patterns.

\section{Conclusions}

This work presented a novel approach to texture recognition using a local des\-criptor derived from techniques of chaos theory. More specifically, we applied a chaotic map to a three-dimensional representation of the image and reconverted that transformed structure back to the image space. Therefore we collected local binary descriptors from these transformed images to compose the texture descriptors.

The methodology was evaluated in the classification of benchmark texture databases and in the identification of Brazilian plant species based on the leaf texture. In both scenarios, the proposed approach demonstrated effectiveness, being competitive with modern solutions presented in the literature, even with some CNN-based methods.

Generally speaking, chaotic maps demonstrated its ability of highlighting non-linear patterns on the image that could not be captured over the original texture. At the same time, a parameter weighting the action of the map in one iteration with that action on the previous iteration guaranteed a smoother transform of the image, attenuating in this way the damaging effect of pattern disorganization typically associated to chaos models. 

Finally, we should mention that our approach does not need to rely specifically on local binary patterns for its local descriptors. In future works, we intend to investigate how other well-known local representations, like bag-of-visual-words, dense-SIFT, etc., could be boosted by chaotic maps in a similar way to that proposed here. 

\section*{Acknowledgements}

This work was supported by the Serrapilheira Institute (grant number Serra-1812-26426). J. B. Florindo also gratefully acknowledges the financial support from National Council for Scientific and Technological Development, Brazil (CNPq) (Grants \#301480/2016-8 and \#423292/2018-8).

\section*{References}

%\bibliographystyle{model2-names}
%\bibliography{chaosMap}

\end{document}